\documentclass[preprint,12pt]{elsarticle}

\usepackage{amssymb}

\usepackage{amsmath,amssymb,amsfonts}
\usepackage{algorithmic}
\usepackage{graphicx}
\usepackage{textcomp}
\usepackage{tabularx}
\usepackage{amsmath}
\usepackage{url}
\usepackage{booktabs}

\journal{Artificial Intelligence in Medicine}

\begin{document}

\begin{frontmatter}



\title{Enhancing Suicide Risk Detection on Social Media through Semi-Supervised Deep Label Smoothing}


%

\author[inst1]{Matthew Squires}

\affiliation[inst1]{organization={School of Mathematics, Physics and Computing, University of Southern Queensland},
            city={ Toowoomba, Australia}}

\author[inst1]{Xiaohui Tao}
\author[inst2]{Soman Elangovan}
\author[inst1]{U Rajendra Acharya}
\author[inst3]{Raj Gururajan}
\author[inst4]{Haoran Xie}
\author[inst3]{Xujuan Zhou}

\affiliation[inst2]{organization={Belmont Private Hospital},
            city={Brisbane, Australia}}


\affiliation[inst3]{organization={School of Business, University of Southern Queensland, Springfield},
            city={Australia}}

\affiliation[inst4]{organization={Department of Computing and Decision Sciences, Lingnan University},
            city={Hong Kong SAR, China}}

\begin{abstract}
Suicide is a prominent issue in society. Unfortunately, many people at risk for suicide do not receive the support required. Barriers to people receiving support include social stigma and lack of access to mental health care. With the popularity of social media, people have turned to online forums, such as Reddit to express their feelings and seek support. This provides the opportunity to support people with the aid of artificial intelligence. Social media posts can be classified, using text classification, to help connect people with professional help. Text classification of social media posts for the detection of mental health distress is a large and ever-expanding field. However, these systems fail to account for the inherent uncertainty in classifying mental health conditions. Unlike other areas of healthcare, mental health conditions have no objective measurements of disease often relying on expert opinion. Thus when formulating deep learning problems involving mental health, using hard, binary labels does not accurately represent the true nature of the data. In these settings, where human experts may disagree, fuzzy or soft labels may be more appropriate. The current work introduces a novel label smoothing method which we use to capture any uncertainty within the data. We test our approach on a five-label multi-class classification problem. We show, our semi-supervised deep label smoothing method improves classification accuracy above the existing state of the art. Where existing research reports an accuracy of 43\% on the Reddit C-SSRS dataset, using empirical experiments to evaluate our novel label smoothing method, we improve upon this existing benchmark to 52\%. These improvements in model performance have the potential to better support those experiencing mental distress. Future work should explore the use of probabilistic methods in both natural language processing and quantifying contributions of both epistemic and aleatoric uncertainty in noisy datasets.
\end{abstract}



\end{frontmatter}

\section{Introduction}
\label{sec:introduction}
Depression and Suicide are significant issues in society. Given this, researchers have explored many avenues to aid in the treatment and diagnosis of these mental health conditions. Barriers such as reduced access to mental health services \cite{Fitzpatrick2021}, and stigma associated with seeking mental health care \cite{Gaur2019} are among the factors which prevent people from seeking help. As such, many people are turning to social media to seek support and share mental health related information \cite{Akhther2022}. The use of Artificial Intelligence (AI) systems to detect depression has seen extensive research. Building upon existing work, Gaur et al. \cite{Gaur2019} investigated the use of text classification to recognise social media users who may be a suicide risk. Text classification systems for suicide behaviours could help to connect users sharing their emotions and mental health struggles with health professionals using AI models.

A difficulty of mental health research, in contrast to other fields of healthcare, mental health conditions have no objective markers of disease \cite{Drysdale2017}. This lack of objective markers is one of several key challenges in identifying psychopathology \cite{Yassin2020}. Furthermore, human raters can find classifying suicidal behaviours using assessment tools to be difficult \cite{Interian2017}. When it is difficult for human raters to agree on a ground truth label it is likely difficult for AI to uncover underlying patterns \cite{Gaur2019}. In turn, this presents a difficulty for deep learning models which have traditionally relied on binary labels. Fuzzy logic, however, allows for an alternative view. When borders between groups are unclear fuzzy variables ``facilitate gradual transitions between states and, consequently, possess a natural capability to express and deal with observation and measurement uncertainties'' \cite[p.4]{Dubois1998}. The inadequacy of binary, ore one hot encoded ground truth labels has also been explored in the field of text emotion classification \cite{Li2023}. Where fuzzy emotions can be used to capture text which may convey multiple emotions, where a binary mapping fails to capture an accurate ground truth. 
\par
Uncertainty is ever present in the mental health field due to the reliance on self-reporting and observation. When uncertainty is present in the labelling of ground truth, it seems unreasonable to use traditional hard labels where $y \in {0,1}$.  In these settings binary variables as ground truth labels do not represent the true nature of a system. Label smoothing is a technique which involves subtracting a small value from the true class, and distributing the subtracted value evenly across each remaining class. Label smoothing is uniform, that is the distribution remains constant across all remaining labels, as below:

\begin{equation}\label{eq:labelSmoothing}
y=
\begin{cases}
1 - \alpha & \text{if $y_i$ = 0} \\
\alpha/(k-1) & \text{otherwise}\\
\end{cases}  
\end{equation}

These updated labels are referred to as soft labels. Initially proposed as a regularisation technique to help prevent model overfitting some research investigates the use of soft labels to improve model performance. Recently, Zhang et al. \cite{Zhang2021} showed non-uniform label smoothing to improve model performance on benchmark datasets such as CIFAR-100 and ImageNet. Our work extends the use of non-uniform distribution label smoothing to text classification and introduces a novel strategy using Bayesian techniques to generate the smoothed labels. These smoothed labels are likely more representative of the fuzzy nature of classifications in the mental health space. As such, this paper seeks to explore issues of uncertainty central to the use of AI in mental health care. Formalised in the following research questions:

\begin{itemize}
\item How can the uncertainty in ground truth labels be expressed in settings where expert opinion may be divided?
\item  How does incorporating uncertainty into labels impact model performance?
\item Can the underlying truth label distribution be found?

\end{itemize}

To explore these research questions we utilise the Reddit C-SSRS dataset first presented by \cite{Gaur2019}. The data includes posts by 500 Reddit users assessed by experts using the Columbia Suicide Severity Rating Scale (C-SSRS).  According to Gaur et al., \cite{Gaur2019} user posts were labelled by four practising clinical psychiatrists with pairwise annotator agreement varying between $\approx 80\%$ and $\approx 60\%$, however, one-hot encoded variables fail to capture the uncertainty between raters. 


Our work is motivated by the idea that it is difficult for deep learning models to identify underlying relationships in data when the labels do not represent the true nature of the system. Thus if human mental health care experts do disagree on how to classify a post then this uncertainty must be expressed. As such, we propose a novel label smoothing method which builds upon existing work to more accurately capture the uncertainty of ground truth labels. Through our experiments exploring the detection of mental distress of social media posts we make the following contributions:

\begin{itemize}
    \item A novel fuzzy semi supervised deep learning method for label smoothing to represent uncertainties in ground truth labels
    \item A state of the art text classification model, achieving accuracy surpassing existing models tested on the same dataset using fuzzy labels;
    \item An exploration of the use of fuzzy class membership for the use of text classification
\end{itemize}

This paper is structured as follows. Section \ref{sec:RelatedWork} provides an overview of existing methods used for uncertainty quantification. These techniques, predominantly used for image segmentation, and the dearth of literature in mental health care motivates this work. This section identifies the gap in the literature and the opportunity for the use of uncertainty estimation in text classification. Section \ref{sec:Methods} provides an overview of the methods, techniques and data set used to present our novel uncertainty estimation techniques incorporated with text classification. In the remaining sections, Section \ref{sec:Results} provides a summary of the performance of the current work compared against the baseline model. Finally, Section \ref{sec:Discussion} and \ref{sec:Conclusion} provide concluding remarks explaining the models behaviour and offering future directions for the field. 

\section{Related Work}\label{sec:RelatedWork}

The quantification of uncertainty when using deep learning in healthcare is expanding rapidly. This is due to the acknowledgement that if deep learning is to be used in critical settings, such as healthcare, uncertainty quantification must be further developed \cite{Begoli2019}. Probabilistic deep learning methods, such as Bayesian Neural Networks, Deep Ensembles and Monte-Carlo (MC) dropout are common techniques proposed to explore these uncertainties. To date, the use of stochastic methods in healthcare focuses on medical imaging and image processing (see \cite{Abdullah2022}). For example, Bayesian techniques were applied to the detection of oral cancer from intraoral images, the diagnosis of COVID-19 from X-rays and the classification of brain lesions from MRI images. 

In their survey, Abdar et al. \cite{Abdar2021} identify aleatoric and epistemic uncertainty as the two main categories of uncertainty. Epistemic uncertainty refers to uncertainty resulting from a model's lack of understanding. H\"{u}llermeir and Waegeman \cite{Huellermeier2021} define epistemic uncertainty as "the uncertainty caused by a lack of knowledge." In the practical sense for AI or deep learning models H\"{u}llermeir and Waegeman \cite{Huellermeier2021} assert epistemic uncertainty "refers to the ignorance of the agent or decision maker." In this context, the "agent or decision maker" could refer to an AI agent or deep learning model. The understanding that epistemic uncertainty refers to the lack of knowledge by a model, has resulted in epistemic uncertainty to be more commonly referred to as model uncertainty \cite{Kendall2017}. Model uncertainty is taken to occur when a model is exposed to an example which lies outside of the distribution on which it was trained. Given this, the logical solution to handle epistemic uncertainty is to provide more data \cite{Kendall2017}. While this may be possible for active learners in fields with large and ever-expanding data sets, in some cases, such as psychology and psychiatry large data sets are not always accessible, making the expression of epistemic uncertainty hugely important. 

In contrast, aleatoric uncertainty is defined as "noise inherent in the data distribution" \cite{Liu2022}. For example, annotation ambiguity. Where annotation ambiguity refers to uncertainty regarding labelling examples in supervised learning tasks. Put another way, H\"{u}llermeir and Waegeman \cite{Huellermeier2021} contend aleatoric uncertainty refers to the randomness inherent to data collection. For this reason, aleatoric uncertainty is also known as data uncertainty. The notion of inherent randomness in data collection is of particular significance in the field of psychiatry and psychology. For many conditions within the mental health space, few or no objective biomarkers of disease exist. As such, the discipline relies heavily on subjective measurement to quantify disease. For example, expert psychiatrists Gaur et al. \cite{Gaur2019} labelled posts on Reddit according to a five-label suicide severity rating scale. Four practising clinical psychiatrists annotated each post according to the five-label classification scheme. Results reported in \cite{Gaur2019} show that pairwise annotator agreement on the annotation of posts varied from 79\& to 65\%. This irreducible randomness due to the subjective nature of mental health conditions is an important consideration when modelling mental health conditions. This disagreement between annotators motivates our work, to explore the utility of fuzzy labels and label smoothing as a more accurate representation of the true data distribution. 

To date, few methods have been explored to represent the subjectiveness of human raters in generating labels for mental health conditions. Data cleaning methods such as label correction were trialled in \cite{Haque2021}, which showed using clustering to correct potentially noisy labels improved model performance. Given the success of label correction, our work seeks to explore the possibility of soft labels to represent uncertainty within the labels. These works move beyond label correction, to label smoothing. 

Techniques for capturing this uncertainty can be divided into Bayesian techniques and deep ensembles. Song et al. \cite{Song2021} applied a Bayesian Deep Neural Network to the classification of oral cancer from a large image dataset. The work utilised Monte Carlo Dropout a common Bayesian technique to express prediction uncertainty. Dropout is a common regularisation technique used to reduce the chances of model overfitting. The technique involves 'dropping' a percentage of nodes from a hidden layer \cite{Caldeira2020}. This dropping is achieved by turning the parameter weights of all edges to a given node to zero, to in essence eliminate the impact of that node on the network. The dropout rate indicates the percentage of nodes from a layer which should be dropped. For example, in Song et al. \cite{Song2021} the dropout rate was set to 0.5 or 50\% of nodes. Seminal work by Gal and Ghahramani \cite{Gal2015} showed combining Monte Carlo simulation with dropout could be used to quantify the uncertainty in model predictions. As hidden layer nodes are dropped randomly, repeatedly simulating the same network with dropout has the effect of in essence training different networks and ultimately leading to more robust networks. The output predictions of each network can then be averaged \cite{Gal2015} and the variance of these predictions is defined as the prediction uncertainty \cite{Song2021}. Song et al. \cite{Song2021} show, the performance of their Convolution Neural Network (CNN) improves when the uncertainty threshold is varied. As such, the aim of uncertainty quantification is that images with high levels of uncertainty can be referred to human experts for further evaluation. 

Additionally, Gour and Jain \cite{Gour2022} provide a second example of the combination of MC dropout with a CNN. Gour and Jain \cite{Gour2022} utilise a CNN for the diagnosis of COVID-19 from x-ray images. The work utilises a dropout rate of 0.425 with a pre-trained CNN. Furthermore, \cite{Wu2022} utilised MC Dropout for the classification of brain lesions from MRI scans. Wu et al. \cite{Wu2022} trained a Deep Convolutional Neural Network (DCNN) with MC Dropout in a teacher-student framework. The teacher-student framework involves the training of two networks, the first model, the teacher is initially trained on the dataset. The second model, the student uses the predictions made from the teacher model with the initial data to make final predictions \cite{Gjestang2021}. Although not Bayesian, Gjestang et al. \cite{Gjestang2021} provide an example of the use of a teacher-student framework for the classification of gastrointestinal images. In Gjestang et al. \cite{Gjestang2021}, the teacher model is first trained on unlabeled data to generate pseudo labels. The student is then trained using the teacher-generated labels, the original data set and the original labels with the aim of minimizing the loss function against the target labels. In a novel approach by Wu et al. \cite{Wu2022} the teacher is a Bayesian Deep Network. The student then receives as inputs the Bayesian probabilities output by the Teacher. However, Wu et al. \cite{Wu2022} do not provide details on the dropout rate of their network. 

A recent review by Abdullah et al. \cite{Abdullah2022} provides a comprehensive overview of the current state of Bayesian deep learning techniques within the field. There exists limited research which incorporates uncertainty quantification and model confidence into natural language processing and text classification problems. Abdullah et al. assert \cite{Abdullah2022} they "are not aware of any published work on medical Natural Language Processing that has used Bayesian deep learning (p.36522, \cite{Abdullah2022}). This lack of published works exploring probabilistic methods, uncertainty quantification and text classification provides the gap which further motivates this work. Additionally, significant calls exist for the addition of uncertainty estimation to deep learning models \cite{Begoli2019}. Given the lack of research exploring uncertainty estimation and text classification and the acknowledged need for deep learning models to express their prediction confidence. Our work seeks to address the needs of the research community by exploring both model and label uncertainty to produce better-performing models, ultimately leading to improved mental health outcomes.

\section{Method}\label{sec:Methods}

\subsection{Conceptual Model}
\begin{figure*}
    \centering
    \includegraphics[scale=0.3]{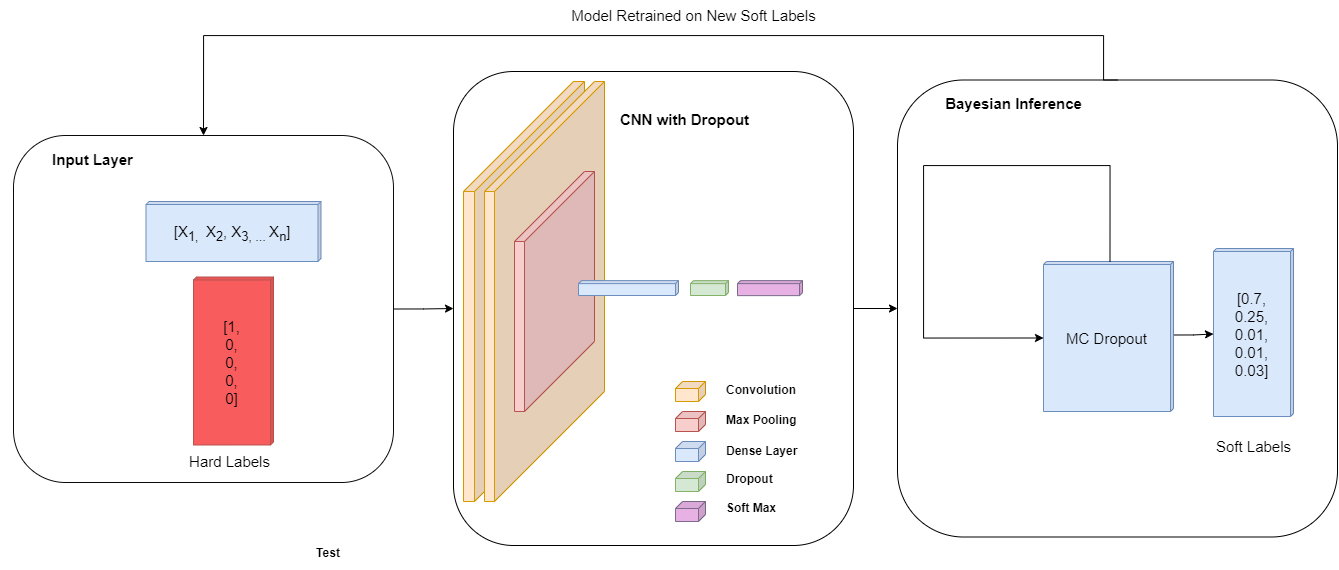}
    \caption{Conceptual Model}
    \label{fig:conceptual_model}
\end{figure*}

This section provides an overview of our novel semi-supervised label smoothing method and the experiments used to evaluate its efficiency when compared to existing methods. As part of the current work, we explore both aleatoric and epistemic uncertainty associated with classifying social media posts and their suicide risk. For the purpose of this work, aleatoric uncertainty can be thought of as the uncertainty associated with the ground truth labels.

Our first experiment explores the label uncertainty which is generated by the subjective nature of the diagnosis of mental health conditions. Given this subjectiveness, human raters can at times disagree on the assessment of the same social media post. To capture this uncertainty we propose fuzzy smoothed labels, generated by our novel label smoothing approach using Bayesian techniques. Additionally, we investigate the impact of prediction confidence on 

This section details the mechanisms of our approach and experiments utilised to measure the efficacy of our model. 

\subsection{Problem Definition}
The current work can be formally defined as a text classification problem. We utilise a dataset, $\mathcal{D}$ which consists of posts from $n$ Reddit users, such that $\mathcal{X}=\{x_1,x_2,x_3\cdots, x_{n}\}$. Users' posts are labelled according to a five-label suicide classification scheme. Such that a post from a user, $x_n$ belongs to a single class $\mathcal{K}=\{k_1,k_2,k_3...k_5\}$. 

Our work explores the impact of using smoothed labels such that an example can belong partially to several classes. For example, the target label of a post $x_n$ could partially belong to multiple classes as demonstrated in equation \ref{eq:labelSmoothing}. In the current work, we introduce a novel method of obtaining the probability of class membership and explore the impact of uniform and non-uniform label smoothing methods against existing methods on the same dataset.

Where our aim is the classifier function which minimises the cost function $C$ during model testing on a hold-out-test-set. Where $C$ is categorical cross-entropy, the multiclass case of the widely used binary cross-entropy, is defined as:

\begin{equation}
    \mathcal{C}=-\sum^k_{k=1}y_k\cdot log(\hat{y_k})
\end{equation}

\subsection{Conceptual Model}

The curated data set provided by Guar et al. \cite{Gaur2019} contains posts from 500 Reddit users labelled according to a five-label suicide severity risk. To perform text classification, natural language must be made machine-readable.

Before text can be used in machine learning tasks it must be prepared for input into deep learning models. All associated preprocessing and model building for this project was constructed in Python with the aid of the software package Keras \cite{Chollet2015}, a popular deep learning library. Preparing natural language text for input to deep learning models requires word embeddings. Text embeddings are the process of converting data to numerical representations \cite{Raaijmakers2022}. A variety of word embedding techniques are available to machine learning researchers to convert Reddit posts to input vectors. Following tokenisation, the input layer takes as input sequential arrays of length 5,041 which is passed to the word embeddings layer. The word embeddings layer learns the most effective vector representation of the model inputs to perform the text classification task.

Figure \ref{fig:conceptual_model} details the proposed model architecture. Using this architecture. The word embedding vector is then passed to the first convolutional layer for model training. The convolutional layer is passed to a second convolutional layer which is then down sampled via a pooling layer. The technical benefits of combining convolutional and max pooling layers for text classification are discussed in Section \ref{sec:CNN}.

Following the max pooling later, the output is flattened before being passed to the final fully connected layer. The fully connected layer utilises the softmax activation function with five output nodes. In our proposed approach Bayesian Inference is used to obtain updated smoothed labels which are then used to retrain the network. Our experiments explore the effectiveness of this approach to represent the uncertainty faced when diagnosing mental health conditions when human experts disagree on a classification. The aim of this work is to capture these disagreements in the data to build a more effective model. 
%
\subsubsection{Convolutional Neural Networks}\label{sec:CNN}
An extension of the multilayer perceptron is the convolutional neural network (CNN) and deep convolutional neural network. Convolutional Neural networks are the workhorse of pattern recognition for images \cite{OShea2015}. The input vector of a multilayer perceptron is typically a flat one-dimensional array or vector. In contrast, the CNN is equipped to take multi-dimensional vectors as inputs. This allows CNNs to take as input a three-dimensional array associated with the red, green and blue components of a two-dimensional image. 

The core components of the CNN which differ from the Multilayer Perceptron (MLP) are the convolutional and max pooling layers. The convolutional layer is known to ``learn local patterns—in the case of images, patterns found in small 2D windows of the inputs" \cite{Chollet2021}. This concept of spacial dimensionality is not possible for flattened inputs to MLPs which instead are constrained to pattern recognition across the entire input vector. The size of the area the convolutional layer looks to for these local patterns is referred to as the kernel \cite{OShea2015}.

In addition to excelling in computer vision tasks, CNNs have demonstrated high levels of performance in text classification \cite{Huang2022}. As with images, this success is attributed to the ability of convolutional layers to learn local patterns within data. In the task of pattern recognition in text, convolutions are able to capture patterns within sequential text. Goldberg \cite{Goldberg2015} asserts the combination of one-dimension convolution and pooling operates as an n-gram detector. With \cite{Jacovi2018} going further, suggesting pooling acts similarly to a feature extraction layer, with only class discriminative n-grams passed through the pooling layer. A pooling layer can be thought of as downsampling the size of the original input \cite{OShea2015}, this ensures relevant information is captured for text classification.

\subsubsection{Fuzzy Label Smoothing}
Label smoothing is a regularisation technique that has been shown to improve model performance. The smoothing of hard labels involves removing a small value from the true case and distributing that value across all classes. A formal definition of uniform label smoothing as described by Shen et al. \cite{Shen2021} is given below:
\begin{equation}
    y_i=
\begin{cases}
1 - \alpha & \text{if $y_i$ = 0} \\
\alpha/(k-1) & \text{otherwise}\\
\end{cases}
\end{equation}

In contrast to existing methods which either use a uniform distribution or denoise data through label correction, our fuzzy label smoothing method allows for partial multi-class membership. Figure \ref{fig:conceptual_model}, details our architecture. We use the initial hard labels in the first training batch before repeatedly simulations using MC Dropout. These simulations produce updated non-uniform smoothed labels. These smoothed labels follow a non-uniform distribution which looks to simulate the uncertainty of ground truth labels. More accurately representing the disagreement in human raters to more accurately represent the underlying distribution.

\subsection{Bayesian Inference and MC Dropout}\label{sec:probability}

Deep learning models have demonstrated human-like or exceeded human performance on many tasks. The concept of the artificial neural network is not new. The foundations for the artificial neuron found their origins in the 1950s. The building block of the artificial network, the perceptron was founded by Rosenblatt et al. \cite{Rosenblatt1958}, and their seminal work on the perceptron. Advances in computing power in the 2010s provided the opportunity for deep neural networks \cite{Duerr2020}. Advances in computing power, have allowed for the chaining of perceptrons together in multiple layers providing the 'depth' of the deep neural network.
Building upon traditional neural networks, Bayesian models have been shown to capture prediction confidence. One popular Bayesian technique is MC dropout, a technique first described in \cite{Gal2015}. 

The following section emphasises the differences between the functionality of a traditional artificial neural network and one which utilises the MC Dropout method. Consider the traditional supervised learning task:

\begin{equation}
    \hat{y} = f(x^{\prime})
\end{equation}

In this task, we seek a function $f$ which minimises the cost function $C$. 

\begin{equation}
    \arg\min C[f]
\end{equation}

Thus a simple feed-forward neural network, with sigmoid activation $\theta$ can be denoted as: 

\begin{equation}
    f(x_{i}) = \theta (x_{i} \cdot w + b)
\end{equation}

As we look to model more complex relationships we can increase the width and depth of the network. Width refers to the number of nodes in a given layer whereas depth refers to the number of hidden layers. Hence deep learning refers to networks with several hidden layers. With these additional hidden layers, the weight term above is replaced with a set of weights $W$ connecting each input node with the hidden layer and a further set of weights connecting each node of the hidden layer with the output node. This set of weights is a weight constellation \cite{Duerr2020}.

Before exploring the use of Monte Carlo Dropout as a Bayesian technique we first provide familiarisation with Bayesian Neural Networks and Bayes theorem. Bayes theorem is shown in Equation \ref{eq:bayes}.

\begin{equation}
\label{eq:bayes}
P(\theta|\mathcal{D}) =  \frac{P(\mathcal{D}|\theta)P(\theta)}{P(\mathcal{D})} 
\end{equation}
The regularisation technique dropout essentially involves turning a proportion of parameter weights down to zero. Visually we can represent this by dropping edges between nodes. When Monte Carlo dropout is used, multiple passes through the network are simulated iteratively reducing the weights of a proportion of nodes from a hidden layer down to zero. When dropout is applied during testing each pass through the network will result in a different weight constellation $W$. Thus for a simulation of $T=100$, we get a set of 100 weights and 100 output predictions. Equation \ref{eq:MCDropout} can be interpreted as the probability of $y$, given the input value $x_i$ from the data set $\mathcal{D}$ is equal to the average of the probabilities for the $x_{i}$th example from each weight constellation, where $\mathcal{W} = (W_{1}, W_{2}, W_{3}\dotsc W_{t})$.

\begin{equation}\label{eq:MCDropout}
    P(y|x_{i}, \mathcal{D}) = \frac{1}{T}\sum_{T}^{t=1}P(y|x_{i}, w_{t})
\end{equation}

Algorithmically we can describe the MC Dropout process as a Monte Carlo simulation through $T$ passes over a test sample. The final output of the Monte Carlo simulation is a probability distribution of membership to each class where the maximum is defined as the final output. 

Hence, for a multiclass problem with k, classes, we can compute the probability of class membership for each weight matrix $w_t$. Such that. The probability of class membership, $p_K$ is given by Equation: \ref{eq:probabilityK}.
\begin{equation}
\label{eq:probabilityK}
    p_k = \frac{1}{T} \sum_{t=1}^{T}p_{k_t} 
\end{equation}

\section{Experiments}
\subsection{Experiment Settings}
\subsubsection{Label Smoothing}
Our computational experiments explore the effects of hard labels, traditional uniform label smoothing and our novel fuzzy label smoothing methods on model performance. To evaluate these methods data was divided into training and test set. In line with the proportions used in \cite{Gaur2019} 20\% of examples were used as a hold-out-test-set. For each experiment, the same convolutional model was used with only the label type modified. The experiment conditions were the original hard labels, uniform label smoothing and finally our novel approach. The results reported are the performance on the test set.

\subsection{Baseline Models}
The baseline model used for the current work is the model presented in \cite{Gaur2019}. Their work provides the existing state-of-the-art performance on the current dataset. The dataset used for this study is sourced from Gaur et al. \cite{Gaur2019}. Gaur et al. \cite{Gaur2019} provide important work on the assessment of suicide risk. The results of which are described later in Section \ref{sec:Results}. Their work provides a novel five-label classification scheme of suicide risk. The five-label scheme classifies Reddit posts from prominent mental health subreddits as either Suicidal Ideation (ID), Suicidal Behaviour (SB), Actual Attempt (AT), Suicide Indicator (IN) or Supportive (SU). Descriptive statistics for the frequency of class membership are described in Table \ref{tab:frequency}.

\begin{table}[]
\caption{Frequency of Class Membership}
    \label{tab:frequency}
    \centering
    \begin{tabular}{c|c}
    \hline
    Label  & Frequency  \\
    \hline
    Ideation   & 171 \\
    \hline
    Behaviour & 77\\
    \hline
    Attempt & 45 \\
    \hline
    Indicator & 99 \\
    \hline
    Supportive & 108 \\
    \hline
    \end{tabular}

\end{table}

\subsection{Metrics}
Common metrics used for binary classification tasks include Accuracy, Precision, Recall and F1-score. However, the current problem is a multi-class problem where the number of classes, $k$, exceeds 2 (ie. $k>2$) as in the binary sense (where $k=2$). Although this paper explores a multi-class problem, common binary classification metrics can be used by incorporating some modifications.

For multi-class classification, the accuracy metric remains the same as in binary classification. Correctly labeled examples are divided by the total number of examples in the test set. However, the precision and recall metrics vary slightly from those used for binary classification. 

From Grandini et al. \cite{Grandini2020}:

\[Precision_k = \frac{TP_{k}}{TP_{k}+FP_{k}} \]

\[Recall_k = \frac{TP_{k}}{TP_{k}+FN_{k}}\]

Where $TP$, is the number of correctly labeled examples and $FP$ is the number of false positives. This Precision metric can be calculated by class or can be summarised using either micro or macro average.  

\[Macro Average Precision = \frac{\sum^{k}_{k=1}Precision_k]}{k}\]

or 

\[Micro Average Precision = \frac{\sum^{k}_{k=1} TP_k}{\sum^{k}_{k}Column_k}\]

The challenge of computing summary metrics for multi-class problems is an open problem in data science. Traditionally, a single summary metric, such as F1-score is preferred to capture mode performance. However, the F1-score using Micro averages fails to capture differences in class sizes \cite{Grandini2020}. Takahashi et al. \cite{Takahashi2021} assert ``the inherit drawback of multi-class F1 scores [is] that these scores do not summarize the data appropriately when a large variability exists between classes. "\cite[p.4966]{Takahashi2021}. The current problem includes large variability in classes. As such, along with average Precision and Recall, we include balanced accuracy. Weighted balanced accuracy is the only metric which captures variability in class sizes. Classes are weighted proportional to their class frequency \cite{Grandini2020}.

\[Weighted Balanced Accuracy = \frac{\sum^{k}_{k=1}Recall_k\cdot\frac{1}{w_k}}{k \cdot w}\]

Where $w_k$ is the assigned weight for each class, $k$ and weights are proportional to the frequency of each class within the sample.

\section{Results}\label{sec:Results}
Our experiments investigate the effect of different labelling methods on model performance. Table \ref{tab:Classification Accuracy} provides an overview of the results of our experiments based on the labelling method used. Initially, we see the results reported by Gaur et al. using their approach. We note, that our proposed architecture trained using hard labels slightly outperforms the results reported by Gaur et al. in accuracy, weighted balanced accuracy, macro average precision and macro average recall.

\begin{table}[h]
\caption{Classification Accuracy by Labelling Method}
    \label{tab:Classification Accuracy}
    \centering
    \footnotesize
    \begin{tabularx}{\textwidth}{XXXXX} \\
   \toprule
    Method & Accuracy & Weighted Balanced Accuracy & Macro Average Precision & Macro Average Recall\\
    \midrule
    Gaur et al. \cite{Gaur2019} & 0.4312 & 0.2567 & 0.2903 & 0.2734   \\
   \midrule
    Hard Labels & 0.4451 & 0.3036 & 0.3337 & 0.3036\\
 \midrule
    Label Smoothing $\alpha = 0.1$ & 0.4699 & 0.3698 & \textbf{0.5284} & 0.3698\\
 \midrule
    Label Smoothing $\alpha = 0.05$ & 0.4783 & 0.4226 & 0.4364 & 0.4266  \\
\midrule
    Deep Bayesian Label Smoothing & \textbf{0.5233} & 0.4923 & \textbf{0.4721}   & \textbf{0.4777}\\
    \bottomrule
\end{tabularx}
    
\end{table}

Label smoothing when $\alpha = 0.1$ improves on both Gaur et al. and the original hard labels. This alpha value records the best macro average precision of all tested methods. Uniform label smoothing with $\alpha=0.05$ produces slight improvements in accuracy and weighted balanced accuracy over the previous alpha value however, we see a slight decline in macro average precision. 

Utilising our novel deep Bayesian label smoothing method achieved the best accuracy (52.33\%) and the best weighted balanced accuracy (49.23\%), by a clear margin with the second best weighted balanced accuracy at 42.26\% demonstrating the ability of the non-uniform label smoothing approach to more accurately predict across all classes within the data

In summary, Table \ref{tab:Classification Accuracy} shows steady improvements in classification accuracy depending on the labelling method used. Uniform soft labels report similar levels of classification accuracy. Whereas, our proposed, non-uniform fuzzy label smoothing method, improves upon existing methods to produce a greater weighted balanced average, and recall over existing methods.

\section{Discussion }\label{sec:Discussion}
The overarching aim of this paper was to explore the effect of various labelling methods to predict mental health distress. The difficulty in developing models on systems which rely heavily on subjective measurements, such as mental health conditions, is binary labels to do not accurately capture disagreement between experts. To capture this uncertainty we present Deep Bayesian Label Smoothing, a new method for softening target labels of deep neural networks to improve prediction accuracy. Given many mental health conditions do not have objective markers of disease. This novel approach is designed to incorporate uncertainty into ground truth labels. Which it is hoped will in turn more accurately represent the true nature of a system, thus leading to improvements in model performance.

Our experiments show, that using soft labels generated using non-uniform label smoothing leads to improved performance on a held-out test set. Existing works on this Reddit C-SSRS curated data set was presented by \cite{Gaur2019}. Interestingly, the existing state-of-the-art work overwhelmingly predicts the predominant class of the dataset. That is, 92\% of predictions of the model presented by \cite{Gaur2019} on the test set are made on the most frequent class of the data. The baseline model exceedingly predicts suicidal ideation, and posts displaying supportive behaviours. This model behaviour is exemplified by low values in precision and recall for the remaining classes. These output metrics suggest the model is failing to uncover the underlying function representation, and is instead making predictions probabilistically. In contrast, our proposed model demonstrates more consistent performance across all classes. This consistent performance suggests some underlying patterns leading to inputs being classified in a certain way do exist. Past research has suggested CNNs may act as n-gram detectors. Goldberg \cite{Goldberg2015} contend that convolutional networks that combine convolutions with pooling layers are useful ``when we we expect to find strong local clues regarding class membership, but these clues can appear in different places in the input" \cite[p.3]{Goldberg2015}. In the sense of the current problem, it is possible there exist class discriminant n-grams identified by the CNN model.

A review of model performance highlights clear variations in performance across classes. The baseline model records low precision and recall for the suicide attempts, suicide behaviours and suicide indicators classes. The entropy-filtered models presented in the current works significantly outperform the baseline model, however, still have no true positives for the suicide attempt class. Our reported results suggest there is no class which was consistently classified as a suicide attempt. Thus the in-text relationships required to detect suicide attempts were not learnt by the CNN model. The inability to detect posts labeled as suicide attempts is one limitation of this model. Demonstrating further improvement of suicide risk classification models is required to excede human performance. Given easy-to-classify posts such as \textit{``Its been a year for me since I survived my Suicide attempt"} should clearly be classified as posts describing a suicide attempt. This post is easily classified when inspected however, it is clear models to date have been unable to detect this class of post. To advance this current work, the use of more sophisticated text representation techniques which incorporate context may improve model accuracy. Examples of text representation techniques which represent contexts such as Glove or BERT \cite{Naseem2021}. It appears it is difficult for the current work to interpret the context, that is if a suicide attempt is referring to another user's post or indicative of a user's own suicide dataset.

The network proposed in this work utilises an embedding layer, two convolutional layers and a max pooling layer. Adopting the convolution-pooling architecture of which the benefits are described in depth in \cite{Goldberg2015}. It is possible deploying deeper convolutional networks, which incorporate more hidden layers may benefit performance. However, deeper networks, these deeper networks become computationally expensive to deploy. Typically, probabilistic methods require large amounts of computational resources. As such, it is possible very deep convolutional networks with probabilistic methods may become too computationally large to compute on a single machine in reasonable computation time and hence difficult to deploy.

Additionally, as stated in Section \ref{sec:RelatedWork}, the authors of \cite{Gaur2019} state that each post had varying levels of agreement on each post. However, \cite{Gaur2019} only presents group-level annotator agreement. The level of inter-annotator agreement at a post level, however, is not included in the dataset. Further details regarding the levels of post uncertainty, hence, understanding the true distribution of aleatoric uncertainty would likely benefit the model's performance. Given this, we assume the distribution of agreement, or aleatoric noise within the data set to not be constant. Future work may explore further probabilistic techniques which account for the varying levels of heteroscedastic aleatoric uncertainty throughout the distribution. 

\section{Conclusion}\label{sec:Conclusion}
The current paper presents Deep Bayesian Label Smoothing, a novel method for generating soft labels for measurements with high levels of uncertainty. We show through empirical experiments our proposed method improves model performance when compared to hard labels and uniform label smoothing. Additionally, the current work provides a text classification system for the assessment of suicide behaviours based on a five-label classification scheme. The incorporation of Bayesian uncertainty techniques into the proposed works greatly improves upon the existing state-of-the-art model on the same dataset. Furthermore, our model provides one of the first adaptions of MC Dropout on a medical text classification task with the majority of work to date focusing on computer vision and image segmentation tasks. Future work may benefit by incorporating word representation models equipped to understand language context such as GloVe or BERT to better classify hard-to-detect classes, such as whether a post is indicative of a suicide attempt behaviour. This is especially true in areas where annotator ambiguity is high.

\section*{Acknowledgment}
This work is partially supported by a grant from Cannon Institute and a grant from the Research Grants Council of the Hong Kong Special Administrative Region, China (Project No. LU HLCA/E/301/23). We gratefully acknowledge the support from Belmont Private Hospital team members, especially Ms Mary Williams (CEO), Rachel Stark (Area Manager), Dr Mark Spelman (Psychiatrist), Dr Sean Gills (Psychiatrist), and Dr Tom Moore (Psychiatrist). Without their kind support, this work wouldn't be possible.

 \bibliographystyle{elsarticle-num} 
 \bibliography{PhD.bib}





\end{document}